\documentclass[manuscript]{acmart}

\usepackage{algorithm}
\usepackage{algpseudocode}
\usepackage{amsmath}
\AtBeginDocument{%
  }

\setcopyright{rightsretained}
\acmConference{AutomationXP26 Workshop of the 2026 CHI Conference on Human Factors in Computing Systems, April 14, 2026, Barcelona, Spain}{}{}
\acmDOI{}
\acmISBN{}

\begin{document}

\title{Building Persona-Based Agents On Demand: Tailoring Multi-Agent Workflows to User Needs}

\author{Giuseppe Arbore}
\authornote{Both authors contributed equally to this research.}
\email{giuseppe.arbore@polito.it}
\orcid{0009-0009-0186-6395}
\affiliation{%
  \institution{Politecnico di Torino}
  \city{Torino}
  \country{Italy}
}

\author{Andrea Sillano}
\authornotemark[1]
\email{andrea.sillano@polito.it}
\orcid{0009-0007-2485-4042}
\affiliation{%
  \institution{Politecnico di Torino}
  \city{Torino}
  \country{Italy}
}

\author{Luigi De Russis}
\email{luigi.derussis@polito.it}
\orcid{0000-0001-7647-6652}
\affiliation{%
  \institution{Politecnico di Torino}
  \city{Torino}
  \country{Italy}
}

\begin{abstract}
Recent advances in agentic AI are shifting automation from discrete tools to proactive multi-agent systems that coordinate multi-specialized capabilities behind unified interfaces. However, today's agent systems typically rely on hard-coded agent architectures with fixed roles, coordination patterns, and interaction flows that limit end-user personalization and make adaptation to individual needs and contexts difficult.
Given this limitation, we argue that on-demand persona-based agent generation offers a promising path towards more efficient and contextually appropriate interaction within agentic workflows. By dynamically crafting agents and personas at run-time to match user characteristics, task demands, and workflow context, agentic platforms can move beyond one-size-fits-all configurations. We present a pipeline for on-demand persona generation in agentic platforms, detailing how real-time crafting of AI personas can be systematically integrated within agent systems, aiming to open new possibilities in agentic platform design paradigms.
\end{abstract}



\maketitle

\pagestyle{plain}
\thispagestyle{plain}

\section{Introduction and Background}
Agentic AI is a rapidly evolving paradigm. Recent advances are shifting automation from discrete, user-invoked tools toward proactive multi-agent systems that coordinate multiple specialized capabilities behind unified interfaces, enabling agents to plan, reason, and autonomously execute multi-step workflows end-to-end~\cite{Pol2025GenerativeAA}. These systems are increasingly supported by concrete architectural components (e.g., persistent memory, structured knowledge resources, reflection mechanisms, and feedback loops) as well as orchestration frameworks that manage coordination and division of labor across multiple agents~\cite{Kumar2025BuildingSA}.
However, recent LLM-based multi-agent systems are often engineered as a set of agents with pre-defined roles (``profiles'') and fixed coordination rules (e.g., who communicates with whom, in what order, and through which intermediate artifacts). Prior research identifies \textit{agent profiling} and \textit{communication} as core architectural dimensions of LLM-based multi-agent systems, and surveys a wide range of approaches in which these choices are specified upfront through prompt templates, role descriptions, and preset interaction protocols~\cite{guo2024survey}.
For example, Bo et al.~\cite{bo2024copper} introduce a collaboration framework in which multiple \textit{actor} agents are guided by a shared \textit{reflector} that generates role-conditioned reflections; while effective, the division of labor and the role schema are designed at the system level~\cite{bo2024copper}.
Similarly, Ding et al.~\cite{ding2024seqcomm} hard-wire a two-phase, asynchronous, multi-level communication scheme where higher-level agents decide first and propagate information and actions to lower-level agents, thereby embedding a specific coordination pattern into the pipeline~\cite{ding2024seqcomm}.

While these system achieve remarkable results in executing and completing tasks, their effectiveness in real-world interaction scenario can be limited by the rigid nature of their configuration. One promising approach to improve LLMs' adaptability is the use of AI-generated personas: constructed identities that changes how the model communicates, reasons, and performs across different tasks and contexts. Persona effects are prompt-sensitive, and changes in how a persona is described (e.g., role-adoption framing, sociodemographic priming, or constraint wording) can substantially alter both open- and closed-ended outputs~\cite{lutz-etal-2025-prompt}. At the same time, persona prompting can influence more than just surface form, as it is able to modulate capabilities and strategy selection. For instance, ``role play prompting'' reports consistent gains in zero shot reasoning across multiple benchmarks, suggesting that persona instruction can induce the model towards effective reasoning paths. Recent work in evaluation further shows that writing style and persona condition formulations can systematically shift performances in different tasks~\cite{truong-etal-2025-persona}. This also motivates persona-aware task like information retrieval mechanism, where embedding-based retrieval can exhibit preferences over certain query styles, affecting how the task is performed~\cite{10.1145/3701551.3703514}. Beyond modulating reasoning strategies, personas can provide an interaction-level control mechanism that possibly reduce friction between users and systems, by aligning responses with user expectation or preferences about tone, verbosity, and tone depth. Empirical studies on chatbot interaction show that variation in linguistic style can significantly affect perceived credibility, engagement, and customer satisfaction indicating that how the model speaks is a measurable metric of interaction quality~\cite{10.1145/3487193, jtaer21020051, 10.1145/3295750.3298956}.
Work on personalized interaction with persona- or profile-conditioned generation to adapt assistance to context shows that users benefit from tailored outputs. For example, we can find this benefit in tutoring system that adapt pedagogy over time or writing assistant that personalize generation to match author's style and preferences~\cite{10.1007/978-3-031-84457-7_13, Liu2024, nicolicioiu2025panzadesignanalysisfullylocal}.    

Building on these observation, LLMs persona are not considered anymore as just styling editors, but instead, they can be actually treated as controlled variables inside end-to-end workflows in agentic architectures. PersonaAgent~\cite{zhang2025personaagentlargelanguagemodel} formalizes personalization for agentic LLM by using a user-specific persona prompt as an intermediary between (i) personalized memory (episodic/semantic) and (ii) personalized actions (tool use), and introduces a test-time alignment strategy that optimizes the persona from recent interactions to better match user. 
Evaluations are also emerging for tool-using settings, such as ETAPP that introduces a benchmark designed specifically to measure personalized tool invocation (not only personalized text), enabling agentic evaluation under diverse user profiles~\cite{hao-etal-2025-evaluating}; or Persona-Plug~\cite{liu-etal-2025-llms} which proposes a plug-and-play persona representation learned from user history (a user embedding attached to inputs) to personalize generation without fine-tuning, offering a practical mechanism for injecting stable user preferences into inference-time behavior. 

As agentic AI moves from discrete automation tools to coordinated workflows of specialized components, current systems still encode roles, coordination patterns, and interaction protocols as fixed design-time choices. This design makes it harder to tailor interaction patterns to different users and contexts without revising prompts, role definitions, or orchestration logic. We argue, instead, for run-time persona-conditioned agent generation, where agents (their roles, interaction policies, and tool-use strategies) are synthesized on demand to match the user, their task, and the evolving workflow state. In this framing, persona-based generation can become a mechanism for adaptivity in agentic platforms: different users and contexts can induce different agent configurations and coordination strategies, possibly leading to more natural interaction and also to different performance profiles depending on who is using the system and under what conditions. To this end, this paper presents a pipeline for on-demand persona generation in agentic platforms, detailing how real-time persona crafting can be systematically integrated within agent systems, enabling context-sensitive adaptivity.

\section{Method }
To address the friction that users can experience when dealing with multi-agent system, we propose a on-demand agent generation that is not tied to a fixed schema of agent profiles.  Consequently, we enable the on-demand synthesis of agent personas at run-time, allowing the system to dynamically craft roles, interaction styles, and coordination behaviors in response to user characteristics, task demands, and workflow context. Instead of requiring manual configuration or predefined agent hierarchies, we treat agent personas as generative constructs that are instantiated and adapted continuously throughout the course of an interaction.

\begin{figure*}[htbp!]
\centering
\includegraphics[width=0.8\textwidth]{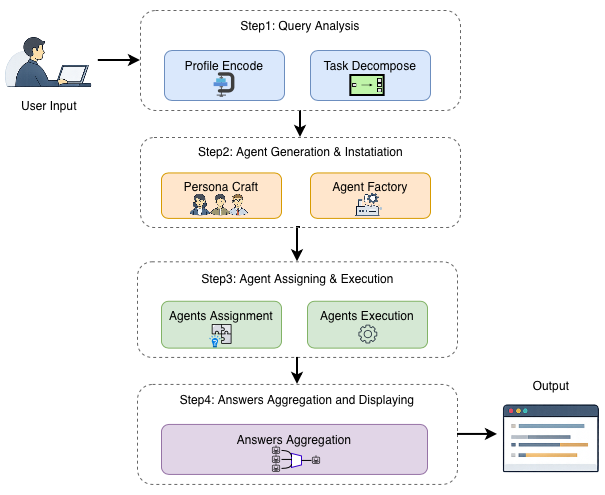}
\caption{A visual representation of the pipeline's steps}
\Description{The image shows the pipeline steps from top to bottom}
\label{fig:steps}
\end{figure*}

The pipeline is initiated when the user submits a query to the system. At this stage, the user interacts with the system freely, without any structural constraints on how the query is formulated as  it can range from a simple question to a complex, multi-part request. This open-ended prompting phase is intentional, as it allows the system to capture the full richness of user intent without forcing predefined templates or interaction patterns. Once the query is received, it is passed to the orchestrator, which serves as the central coordinator of the entire pipeline, managing all subsequent steps without any further intervention required from the user. 
The on-demand persona generation is composed by four different steps in the orchestrator: (i) Query Analysis, (ii) Agent Generation and Instantiation, (iii) Agent Assigning and Execution, and (iv) Answers Aggregation and Displaying.
The pipeline operates on a session-based model, meaning that all the contextual information (e.g., user information, generated persona, and spawned agents) gathered during an interaction are bounded to a single session. This is a deliberate design choice grounded on two main reasons: keeping coherence across the same session with multiple queries, and ensuring that system is not task dependent and can easily adapt itself. An overview of how the pipeline process work is shown in Algorithm~\ref{alg:simplified}, while a step-by-step visualization is represented in Figure~\ref{fig:steps}.

\paragraph{\textbf{Step 1 - Query Analysis}} The first step executed by the orchestrator is responsible for transforming the raw user input into a structured representation that can be later used in the pipeline. It consists of two sub-steps that run in sequence. 
The first sub-step is \textit{ProfileEncode}, which uses implicit or explicit signal from the user query to create a representation of the user and their intent. This profile is intended to capture attributes like domain expertise, preferred communication style, task familiarity, and, crucially, the intent of the query, serving as conditioning knob for persona crafting in the next steps. The second sub-step is \textit{TaskDecompose}, where the orchestrator breaks the query down into a set of discrete tasks. Ideally, this decomposition is more than a flat list; instead, it has to capture dependencies between tasks, determining which tasks can be executed in parallel and which must follow a sequential order. The output of this step is a structured task plan that can be later employed by the orchestrator to manage the generated agents. 

\paragraph{\textbf{Step 2 - Agent Generation and Instantiation}} The second step is responsible of translating the plan produced the previous step into a set of operational agents. The first sub-step is \textit{PersonaCraft}, where the orchestrator can dynamically craft a persona for each sub-task. This step is the central part of the entire pipeline. Instead of relying on predefined role the system can generate needed personas and agents on-demand, allowing for a non-fixed architecture. Moreover, this process is aware of already existing agents and roles, allowing for precise persona development and avoiding duplicates. Concretely, a persona is a structured specification that defines the agent's role, the domain competencies it should exhibit, the communication style it should adopt when contributing to the final response, and what capabilities it owns. All these pieces of information are extracted in the previous step.
The second sub-step is \textit{AgentFactory}, where each synthesized persona is used to initialize one agent instance. The persona specification is passed to configure an LLM-backed agent, shaping its reasoning style, interaction behavior, and output format. The output of this step is a pool of agents with assigned personas, optimized for task-specific goals.

\paragraph{\textbf{Step 3 - Agent Assigning and Execution}} The third step executed in the pipeline coordinates collaboration among agents by assigning each task to its corresponding agent, determining an execution order derived directly from the dependency structure of the tasks produced during decomposition, and managing dependencies and information flow. Tasks with no dependencies  each one other are scheduled for parallel execution, while those that depend on the output of a preceding task are queued for sequential execution, ensuring that no agent begins its task before the information it requires is available.
Each agent executes its assigned task and produces a partial result that is used by the orchestrator that, after inspecting the dependency graph to determine which agents require that result as input, passes it accordingly before triggering their execution. This controlled information routing ensures that each agent operates with the context it needs while remaining decoupled from the internal workings of other agents in the pool, preserving the modularity of the system. 

\paragraph{\textbf{Step 4 - Answers Aggregation and Displaying}} After all agents have produced their results, the system integrates them into a final response. This aggregation typically involves selecting, merging answers, and resolving eventual inconsistencies across agents, removing redundancies, and aligning the final output with the user's requested style and format. The system then delivers a single coherent answer that preserves the benefits of specialization while remaining concise, consistent, and directly usable by the user.

At each new user query within the same session, the pipeline is re-executed. Previously instantiated agents and persona profiles are retained rather than discarded; when additional specialization is required, new agents/personas can be created and appended to the existing pool.

\begin{algorithm}[t]
\caption{On-Demand Persona-Based Agent Generation}
\label{alg:simplified}
\begin{algorithmic}[1]

\Require User queries $q$
\Ensure Response $\mathcal{R}$ delivered to user

\Statex \textbf{Step 1: Query Analysis}
\State Extract user characteristics $u \leftarrow \textsc{ProfileEncode}(\mathcal{U}, q)$
\State Decompose query into tasks: $\{t_1, \ldots, g_n\} \leftarrow \textsc{TaskDecompose}(q, u)$

\Statex \textbf{Step 2: Agent Generation and Instantiation}
\For{each task $t_i$}
    \State Synthesize persona: $p_i \leftarrow \textsc{PersonaCraft}(t_i, u, p)$
    \State Instantiate persona-based agent: $a_i \leftarrow \textsc{AgentFactory}(p_i)$
\EndFor

\Statex \textbf{Step 3: Agents Assigning and Execution}
\State Orchestrator assigns $t_i$ to $a_i$ and manages execution order
\For{each agent $a_i$ in execution order}
    \State Agent execution: $r_i \leftarrow a_i.\textsc{Execute}(t_i)$

\EndFor

\Statex \textbf{Step 4: Answers Aggregation and Displaying}
\State Orchestrator answer aggregation: $\mathcal{R} \leftarrow \textsc{Aggregate}(\{r_1, r_2, \ldots, r_n\})$
\State \Return $\mathcal{R}$ to user

\end{algorithmic}
\end{algorithm}

\section{Conclusions}
This paper outlines how AI-generated personas can be integrated into agentic platforms to move beyond traditional schema-fixed architectures, thereby increasing personalization and adaptability across users and tasks. Our proposal presents a system, where persona-conditioned agent generation is treated as one of the architectural pillars alongside tool use, memory, and orchestration in agentic platforms. By moving agent roles, interaction styles, and behaviors from design-time constants to runtime variables, we aim to make systems being able to adapt and align the agents and the users in real-time. 
Users are no longer forced to understand and to adapt to a fixed agent topology; instead, it is the system that tailors itself to the user. Our proposal can carry implications that may go beyond usability, as it positions on-demand persona-augmented agent generation as a mechanism to empower end-users, enabling them to increase personalized interactions across different backgrounds, expertise levels, and task contexts. As agentic platforms grow in capability and complexity, leveraging persona-aware designs combined with runtime adaptability may contribute to keep these systems accessible, adaptable, and meaningful to users.

\bibliographystyle{ACM-Reference-Format}
\bibliography{bibfile}

\end{document}